\documentclass[10pt,twocolumn,letterpaper]{article}

\usepackage{cvpr}
\usepackage{times}
\usepackage{epsfig}
\usepackage{graphicx}
\usepackage{amsmath}
\usepackage{amssymb}

\usepackage{multirow}
\usepackage{subcaption}
\usepackage{enumitem}

\usepackage[breaklinks=true,bookmarks=false]{hyperref}

\cvprfinalcopy 

\def\cvprPaperID{****} 

\begin{document}

\title{Funnel-Structured Cascade for Multi-View Face Detection with Alignment-Awareness}

\author{Shuzhe Wu$^1$ \and Meina Kan$^{1,2}$ \and Zhenliang He$^1$ \and Shiguang Shan$^{1,2}$ \and Xilin Chen$^1$ \\
$^1$Key Laboratory of Intelligent Information Processing of Chinese Academy of Sciences (CAS)\\ Institute of Computing Technology, CAS, Beijing, 100190, China\\
$^2$CAS Center for Excellence in Brain Science and Intelligence Technology\\
{\tt\small \{shuzhe.wu, meina.kan, zhenliang.he\}@vipl.ict.ac.cn, \{sgshan, xlchen\}@ict.ac.cn}
}

\maketitle

\begin{abstract}
   Multi-view face detection in open environment is a challenging task due to diverse variations of face appearances and shapes. Most multi-view face detectors depend on multiple models and organize them in parallel, pyramid or tree structure, which compromise between the accuracy and time-cost. Aiming at a more favorable multi-view face detector, we propose a novel funnel-structured cascade (FuSt) detection framework. In a coarse-to-fine flavor, our FuSt consists of, from top to bottom, 1) multiple view-specific fast LAB cascade for extremely quick face proposal, 2) multiple coarse MLP cascade for further candidate window verification, and 3) a unified fine MLP cascade with shape-indexed features for accurate face detection. Compared with other structures, on the one hand, the proposed one uses multiple computationally efficient distributed classifiers to propose a small number of candidate windows but with a high recall of multi-view faces. On the other hand, by using a unified MLP cascade to examine proposals of all views in a centralized style, it provides a favorable solution for multi-view face detection with high accuracy and low time-cost. Besides, the FuSt detector is alignment-aware and performs a coarse facial part prediction which is beneficial for subsequent face alignment. Extensive experiments on two challenging datasets, FDDB and AFW, demonstrate the effectiveness of our FuSt detector in both accuracy and speed.
\end{abstract}

\section{Introduction}

Fast and accurate detection of human faces is greatly demanded in various applications. While current detectors can easily detect frontal faces, they become less satisfactory when confronted with complex situations, e.g. to detect faces viewed from various angles, in low resolution, with occlusion, etc. Especially, the multi-view face detection is quite challenging, because faces can be captured almost from any angle - even exceeding $90^{\circ}$ in extreme cases, leading to significant divergence in facial appearances and shapes.

Along with the steady progress of face detection, there have been mainly three categories of face detectors with different highlights. The most classic are those following the boosted cascade framework \cite{yang2014, li2013, chen2014}, originating in the seminal work of Viola and Jones \cite{viola2004}. These detectors are quite computationally efficient, benefited from the attentional cascade and fast feature extraction. Then to explicitly deal with large appearance variations, deformable part models (DPM) \cite{felzenszwalb2010} are introduced to simultaneously model global and local face features \cite{zhu2012, yan2014, mathias2014}, providing an intuitive way to cover intra-class variations and thus being more robust to deformations due to pose, facial expressions, etc. DPM has established a reputation for its promising results on challenging datasets, but detection with DPM is time-consuming, inspiring researches on speeding up techniques \cite{yan2014}. Recently, detectors based on neural networks, e.g. convolutional neural networks (CNN) \cite{farfade2015icmr, li2015cvpr, yang2015iccv:faceness, zhan2015neurocomputing, tao2016neurocomputing, jiang2016neurocomputing}, have attracted much attention and achieved magnificent accuracy on the challenging FDDB dataset \cite{vidit2010tr}, as they enjoy the natural advantage of strong capability in non-linear feature learning. The weakness of CNN-based detectors is their high computational cost due to intensive convolution and complex nonlinear operations.

\begin{figure*}
\begin{center}
    \begin{subfigure}[t]{0.25\linewidth}
        \centering
        \includegraphics[width=\linewidth]{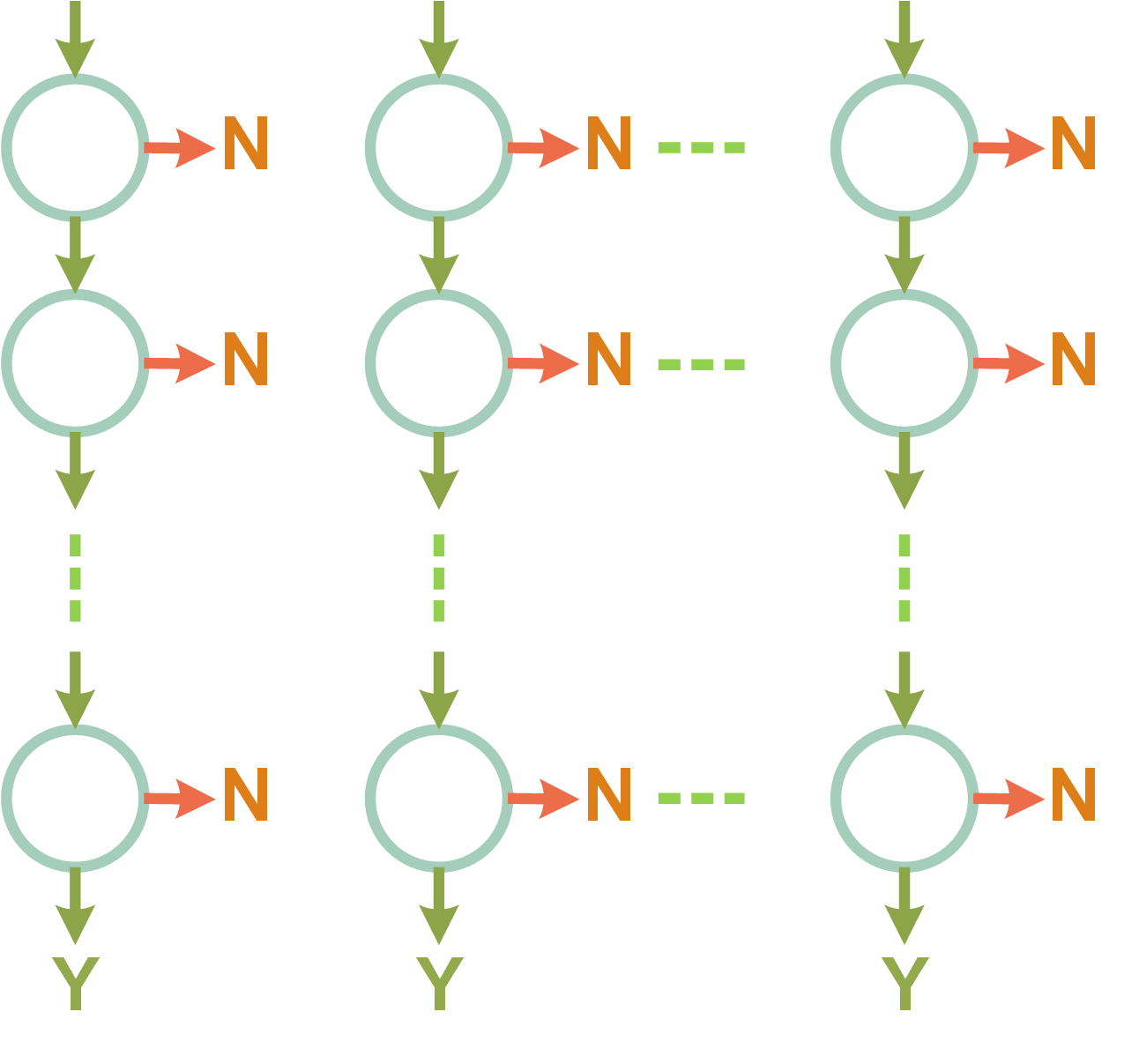}
        \caption{Parallel structure}
        \label{fig:fig1_parallel_structure}
    \end{subfigure}
    ~
    \begin{subfigure}[t]{0.28\linewidth}
        \centering
        \includegraphics[width=\linewidth]{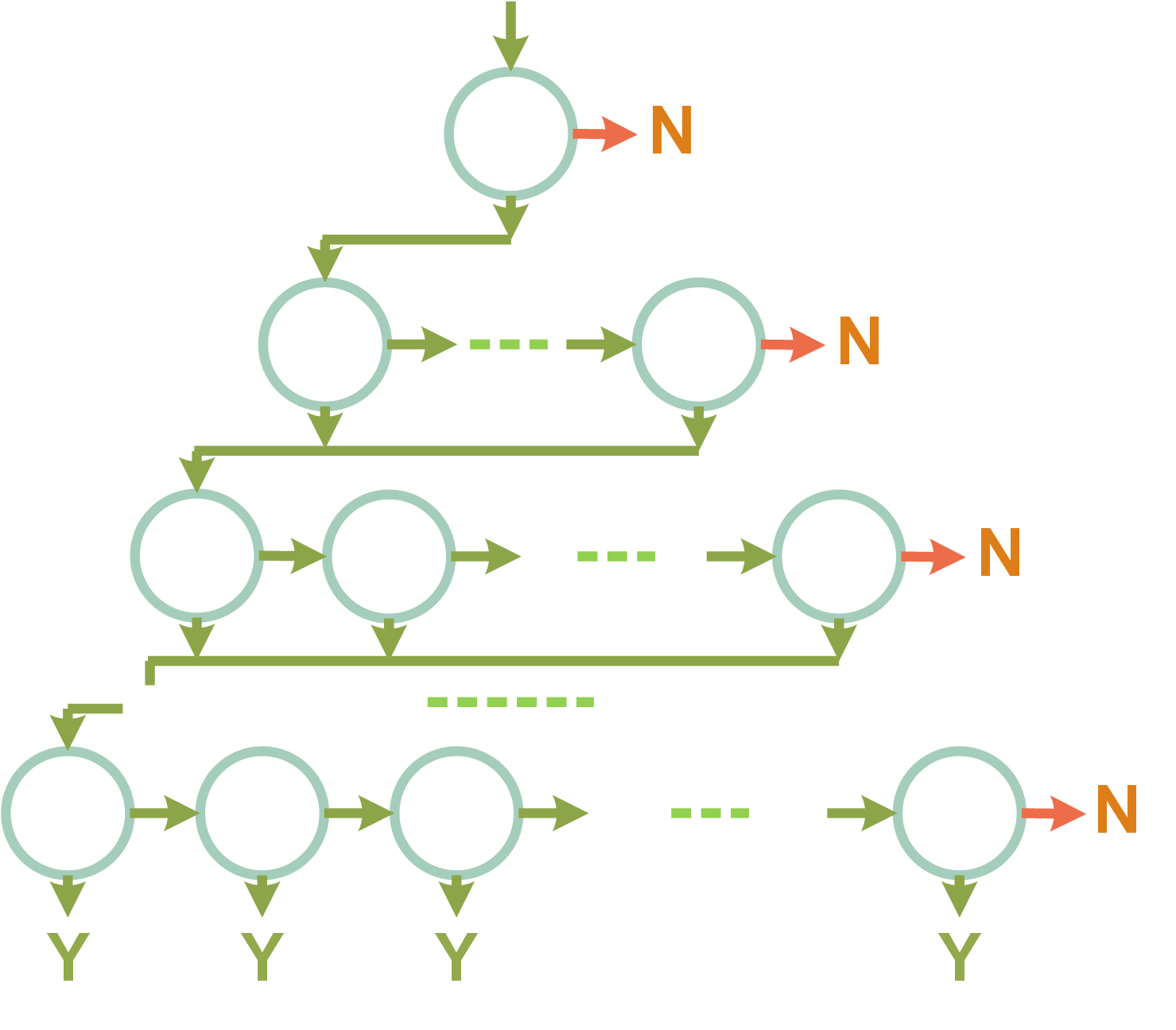}
        \caption{Pyramid structure}
        \label{fig:fig1_pyramid_structure}
    \end{subfigure}
    ~
    \begin{subfigure}[t]{0.28\linewidth}
        \centering
        \includegraphics[width=\linewidth]{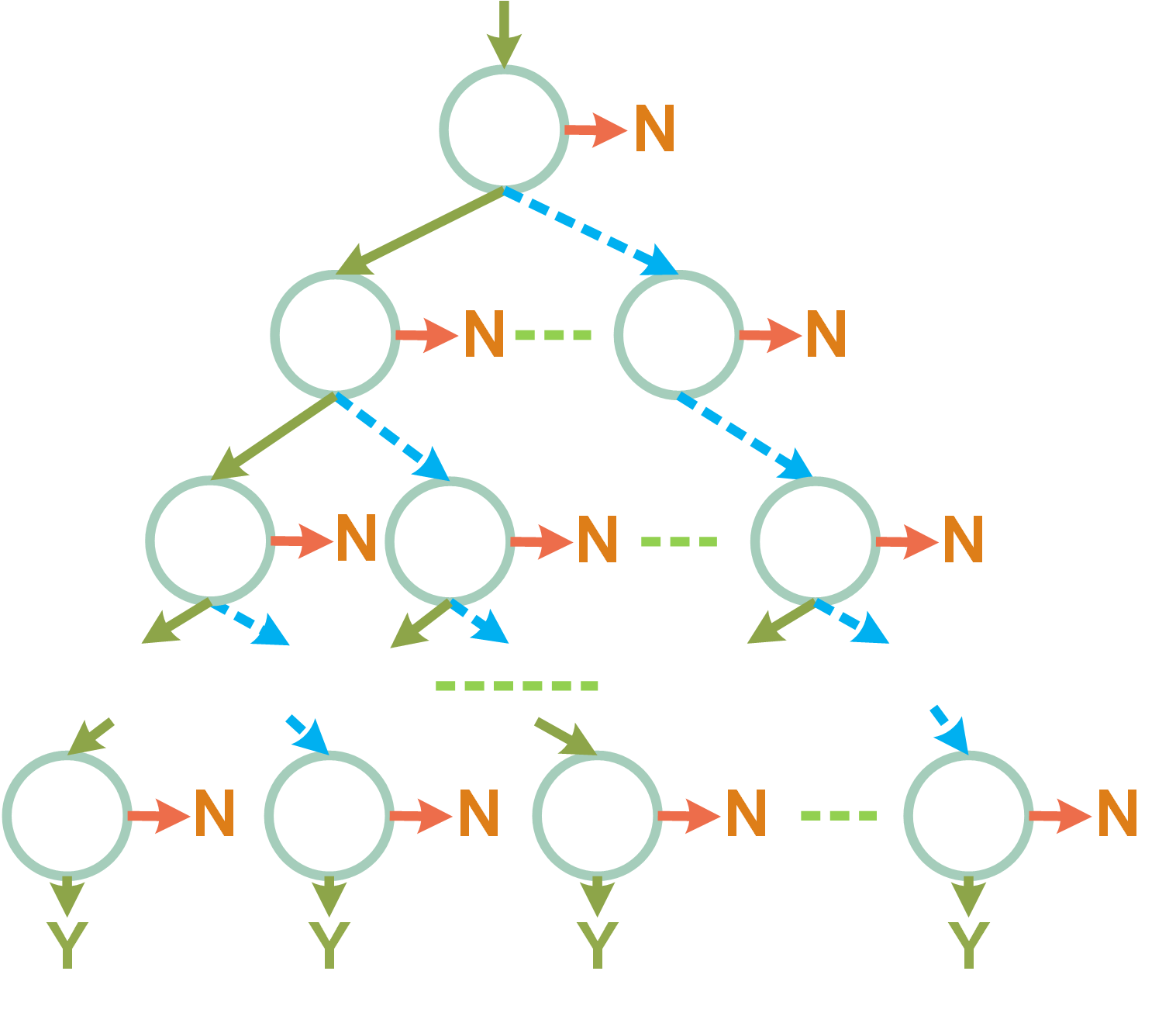}
        \caption{Tree structure}
        \label{fig:fig1_tree_structure}
    \end{subfigure}
\end{center}
   \caption{Different structures for multi-view face detection.}
\label{fig:different_structure}
\end{figure*}

Most works mentioned above focus on designing an effective detector for generic faces without considerations for specific scenarios such as multi-view face detection. In order to handle faces in different views, a straightforward solution is to use multiple face detectors in parallel \cite{li2013, yang2014, mathias2014}, one for each view, as shown in Figure \ref{fig:fig1_parallel_structure}. The parallel structure requires each candidate window to be classified by all models, resulting in an increase of the overall computational cost and false alarm rate. To alleviate this issue, each model needs to be elaborately trained and tuned for better discrimination between face and non-face windows, ensuring faster and more accurate removal of non-face windows.

More efficiently, the multiple models for multi-view face detection can be organized in a pyramid \cite{li2002eccv} or tree structure \cite{huang2007tpami}, as shown in Figure \ref{fig:fig1_pyramid_structure} and \ref{fig:fig1_tree_structure}, forming a coarse-to-fine classification scheme. In such structures, the root classifier performs the binary classification of face vs. non-face, and then at subsequent layers, faces are divided into multiple sub-categories with respect to views in a finer granularity, each of which is handled by an independent model. The pyramid structure is actually a compressed parallel structure with shared nodes in higher layers or a stack of parallel structures with different view partitions. Therefore the pyramid-structured detectors suffer from similar problems that parallel-structured ones are faced with. The tree-structured detectors are different in that branching schemes are adopted to avoid evaluating all classifiers at each layer, but this can easily lead to missing detections with incorrect branching. To relax the dependence on accurate branching, Huang \etal \cite{huang2007tpami} designs a vector boosting algorithm to allow multiple branching.

\begin{figure}
\begin{center}
   \includegraphics[width=0.7\linewidth]{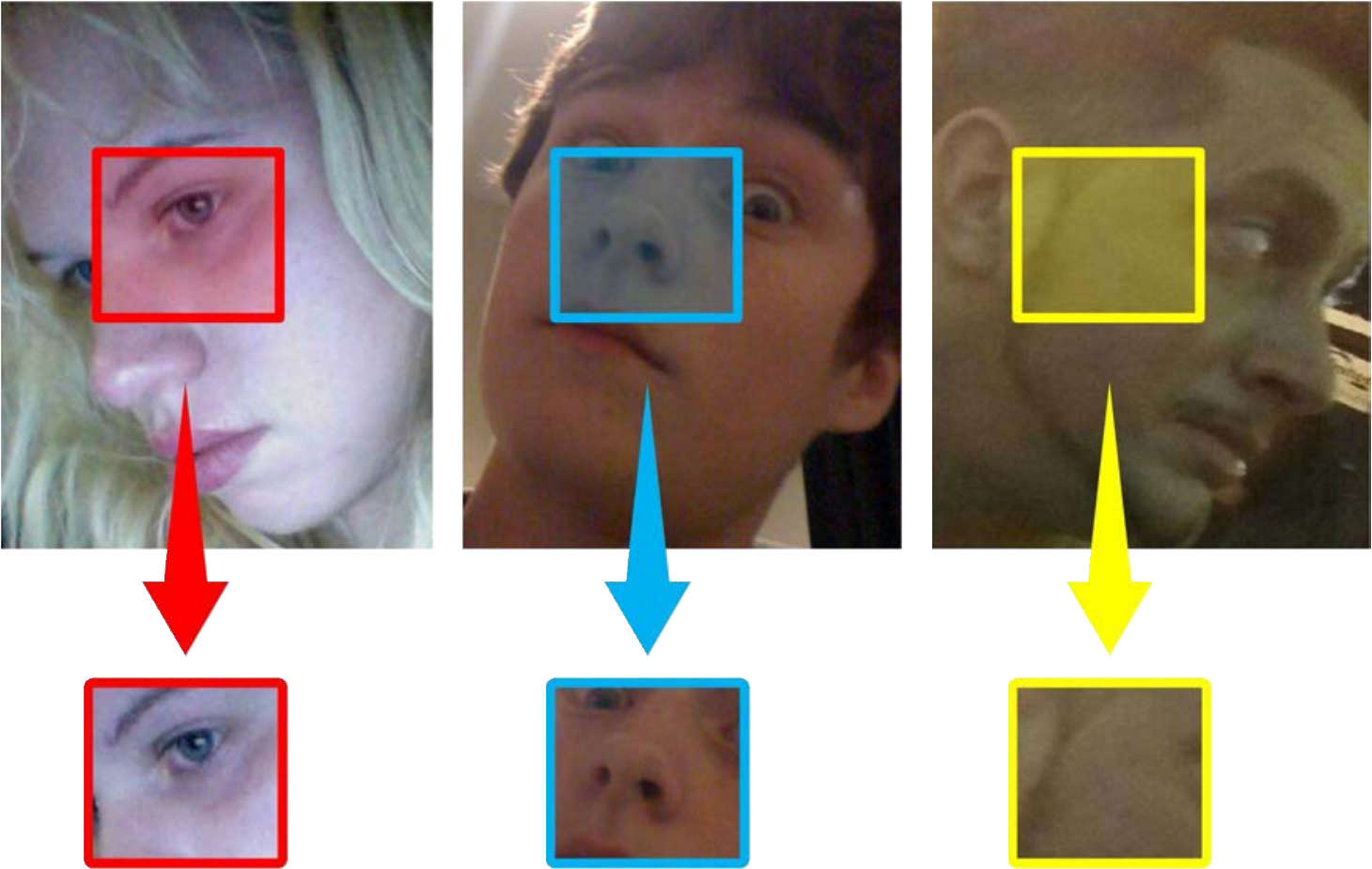}
\end{center}
   \caption{The window at the same position on three faces in varied views contain totally distinct semantics.}
\label{fig:faces_not_aligned}
\end{figure}

Considering the appearance divergence of multi-view faces from the perspective of feature representation, the intra-class variations are mainly due to features extracted at positions with inconsistent semantics. For instance, in Figure \ref{fig:faces_not_aligned}, three faces in different views are shown and the window at the same positions on different faces contains completely distinct semantics, resulting in features describing eye, nose and cheek respectively. Thus there does not exist a good correspondence between representations of faces in different views. Chen \etal \cite{chen2014} compares densely extracted features with shape-indexed features and finds the latter to be more discriminative. By using features at aligned landmarks, faces in different views can be more compactly represented and better distinguished from non-face regions.

\begin{figure*}
\begin{center}
   \includegraphics[width=0.8\linewidth]{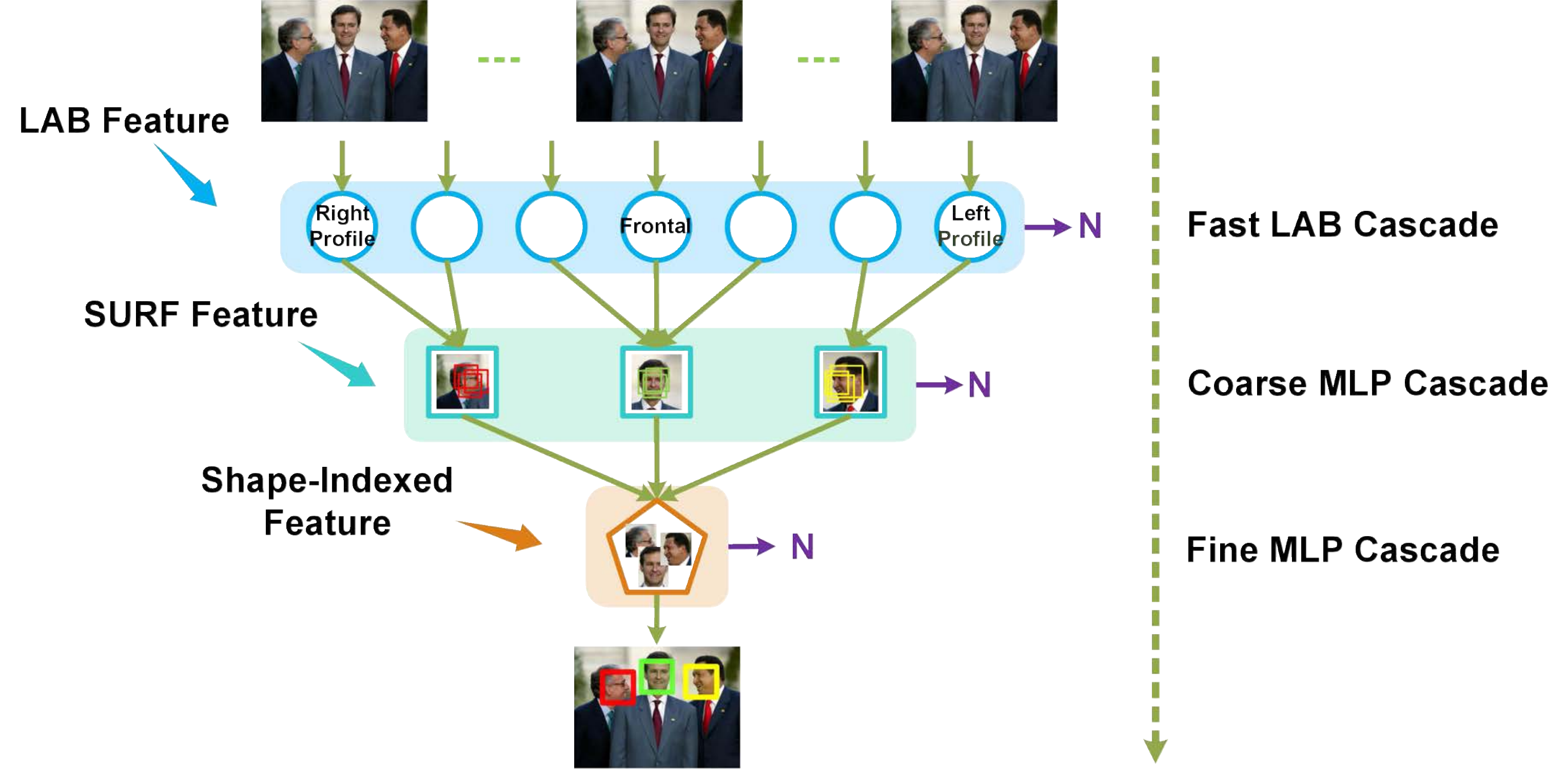}
\end{center}
   \caption{An overview of our proposed funnel-structured cascade framework for multi-view face detection.}
\label{fig:funnel_structure}
\end{figure*}

To provide a more effective framework for multi-view face detection, we design a novel funnel-structured cascade (FuSt) multi-view face detector, which enjoys both high accuracy and fast speed. The FuSt detector, as shown in Figure \ref{fig:funnel_structure}, features a funnel-like structure, being wider on the top and narrower at the bottom, which is evidently different from previous ones. At early stages from the top, multiple fast but coarse classifiers run in parallel to rapidly remove a large proportion of non-face windows. Each of the parallel classifiers is trained specifically for faces within a small range of views, so they are able to ensure a high recall of multi-view faces. By contrast, at subsequent stages, fewer classifiers, which are slightly more time-consuming but with higher discriminative capability, are employed to verify the remaining candidate windows. Gathering the small number of windows surviving from previous stages, at the last stages at the bottom, a unified multilayer perceptron (MLP) cascade with shape-indexed features is leveraged to output the final face detection results. From top to bottom, the number of models used decreases while the model complexity and discriminative capability increase, forming a coarse-to-fine framework for multi-view face detection.

Compared with previous multi-view face detectors, the proposed FuSt detector is superior in that a more effective framework is used to organize multiple models. The contribution of our work compared to existing literature is listed as below.
\begin{itemize}[noitemsep]
\item First, a unified MLP cascade is leveraged as last few stages to examine proposals provided by previous stages, which addresses the problem of increased false alarm rate resulting from using multiple models in other structures, e.g. parallel or tree structure.
\item Second, the proposed FuSt detector operates in a gathering style instead of adopting any branching mechanism as in pyramid- or tree-structured detectors. Therefore it can naturally avoid missing detections caused by incorrect branching and reach a high recall.
\item Third, in the final unified MLP cascade, features are extracted in semantically consistent positions by integrating shape information rather than fixed positions as in conventional face detectors, and thus multi-view faces can be better distinguished from non-face regions. Moreover, the extra shape output from our FuSt detector can provide a good initialization for subsequent alignment.
\item With extensive experiments on challenging face detection datasets including FDDB \cite{vidit2010tr} and AFW \cite{zhu2012}, the FuSt detector is demonstrated to have both good performance and fast speed.
\end{itemize}

The rest of the paper is organized as follows. Section \ref{sec:funcas} describes the proposed FuSt detector in detail, explaining the design of different stages from top to bottom. Section \ref{sec:expt} presents the experimental results on two challenging face detection datasets together with analysis on the structure and shape prediction. The final Section \ref{sec:conclude} concludes the paper and discusses the future work.

\section{Funnel-Structured Cascade Multi-View Face Detector}
\label{sec:funcas}

An overview of the framework of FuSt detector is presented in Figure \ref{fig:funnel_structure}. Specifically, the FuSt detector consists of three coarse-to-fine stages in consideration of both detection accuracy and computational cost, i.e. Fast LAB Cascade classifier, Coarse MLP Cascade classifier, and Fine MLP Cascade classifier. An input image is scanned according to the sliding window paradigm, and each window goes through the detector stage by stage.

The Fast LAB Cascade classifiers aim to quickly remove most non-face windows while retaining a high call of face windows. The following Coarse MLP Cascade classifiers further roughly refine the candidate windows at a low cost. Finally the unified Fine MLP Cascade classifiers accurately determine faces with the expressive shape-indexed features. In addition, it also predicts landmark positions which are beneficial for subsequent alignment.

\subsection{Fast LAB Cascade}

For real-time face detection, the major concern in the sliding window paradigm is the large quantity of candidate windows to be examined. For instance, to detect faces with sizes larger than $20\times 20$ on a $640\times 480$ image, over a million windows need to be examined. Hence it is quite necessary to propose a small number of windows that are most likely to contain faces at minimal time cost.

A good option for fast face proposal is to use boosted cascade classifiers, which are very efficient for face detection task as shown by Viola and Jones \cite{viola2004}. Yan \etal \cite{yan2008cvpr} propose an efficient LAB (Locally Assembled Binary) feature, which only considers the relative relations between Haar features, and can be accelerated with a look-up table. Extracting an LAB feature in a window requires only one memory access, resulting in constant time complexity of $O(1)$. Therefore we employ the more preferable LAB feature with boosted cascade classifiers, leading to the extremely fast LAB cascade classifiers, which are able to rapidly reject a large proportion of non-face windows at the very beginning.

Although the LAB feature is quite computationally efficient, it is less expressive and has difficulty modeling the complicated variations of multi-view faces for a high recall of face windows. Therefore, we adopt a divide-and-conquer strategy by dividing the difficult multi-view face detection problem into multiple easier single-view face detection problems. Specifically, multiple LAB cascade classifiers, one for each view, are leveraged in parallel and the final candidate face windows are the union of surviving windows from all of them.

Formally, denote the whole training set containing multi-view faces as $S$, and it is partitioned into $v$ subsets according to view angles, denoted as $S_i, i = 1, 2, \cdots, v$. With each $S_i$, an LAB cascade classifier $c_i$ is trained, which attempts to detect faces in the $i$-th view angle. For a window $x$ within an input image, whether it is possible to be a face is determined with all LAB cascade classifiers as follows:
\begin{align} \label{eq:layer1}
y = c_1 (x) \vee c_2 (x) \vee \cdots \vee c_v (x),
\end{align}
where $y \in \{0,\,1\}$ and $c_i(x) \in \{0,\,1\}$ indicate whether $x$ is determined to be a face or not. As can be seen from Eq. \eqref{eq:layer1}, a window will be rejected if and only if it is classified as negative by all LAB cascade classifiers. Using multiple models will cost more time, but all models can share the same LAB feature map for feature extraction. Therefore more models add only minor cost and the overall speed is still very fast as a high recall is reached.

Besides the high recall, the parallel structure also allows more flexibility in view partitions. Since it does not suffer from missing detections caused by incorrect branching as in tree structure, a rough rather than an accurate view partition is enough. In other words, degenerated partitions with incorrect view labeling of faces has minor influences on the overall recall of all LAB cascade classifiers. It is even applicable for automatic view partition from clustering or that based on other factors.

\subsection{Coarse MLP Cascade}

After the stages of LAB cascade, most of the non-face windows have been discarded, and the remaining ones are too hard for the simple LAB feature to handle. Therefore, on subsequent stages, the candidate windows are further verified by more sophisticated classifiers, i.e. MLP with SURF (Speeded-Up Robust Feature) \cite{li2011iccvw}. To avoid imposing too much computational cost, small networks are exploited to perform a better but still coarse examination.

SURF features are more expressive than LAB features, but are still computationally efficient benefited from the integral image trick. Therefore face windows can be better differentiated from non-face windows with low time cost. Furthermore, MLP is used with SURF feature for window classification, which can better model the non-linear variations of multi-view faces and diverse non-face patterns with the equipped nonlinear activation functions.

MLP is a type of neural network consisting of an input layer, an output layer, and one or more hidden layers in between. An $n$-layer MLP $F(\cdot)$ can be formulated as
\begin{align}
& F(x) = f_{n - 1} (f_{n-2} (\cdots  f_1(x) )), \label{eq:mlp_function} \\
& f_i(z) = \sigma(W_i z + b_i). \label{eq:sigmoid_function_1}
\end{align}
where $x$ is the input, i.e. the SURF features of a candidate window; $W_i$ and $b_i$ are the weights and biases of connections from layer $i$ to $i + 1$ respectively. The activation function $\sigma(\cdot)$ is commonly designed as a nonlinear function such as a sigmoid function $\sigma(x) = \frac{1}{1 + \mathrm{e}^{-x}}$. As can be seen in Eq. \eqref{eq:mlp_function} and \eqref{eq:sigmoid_function_1}, units in hidden layers and output layer are both equipped with nonlinear functions, so the MLP is endowed with strong capability to model highly nonlinear transformations. The training of MLPs aims to minimize the mean squared error between the predictions and the true labels as below
\begin{align} \label{eq:mlp_loss}
\min_F \sum_{i=1}^n \Vert F(x_i) - y_i \Vert^2 ,
\end{align}
where $x_i$ is the feature vector of the $i$-th training sample and $y_i$ the corresponding label as either $1$ or $0$, representing whether the sample is a face or not.
The problem in Eq. \eqref{eq:mlp_loss} can be easily solved by using gradient descent under the back propagation framework \cite{schmidt2005minfunc}.

We employ multiple coarse MLPs to construct an attentional cascade, in which the number of features used and the size of the network gradually increase stage by stage. The SURF features used at each stage is selected by using group sparse \cite{eldar2010tsp}. Since the MLP cascade classifiers have stronger ability to model face and non-face variations, windows passing through multiple LAB cascade classifiers can be handled together by one model, i.e. one MLP cascade can connect to multiple LAB cascade classifiers.

\subsection{Fine MLP Cascade with shape-indexed feature}
\label{ssec:fine_mlp_cas}

\begin{figure*}
\begin{center}
   \includegraphics[width=0.7\linewidth]{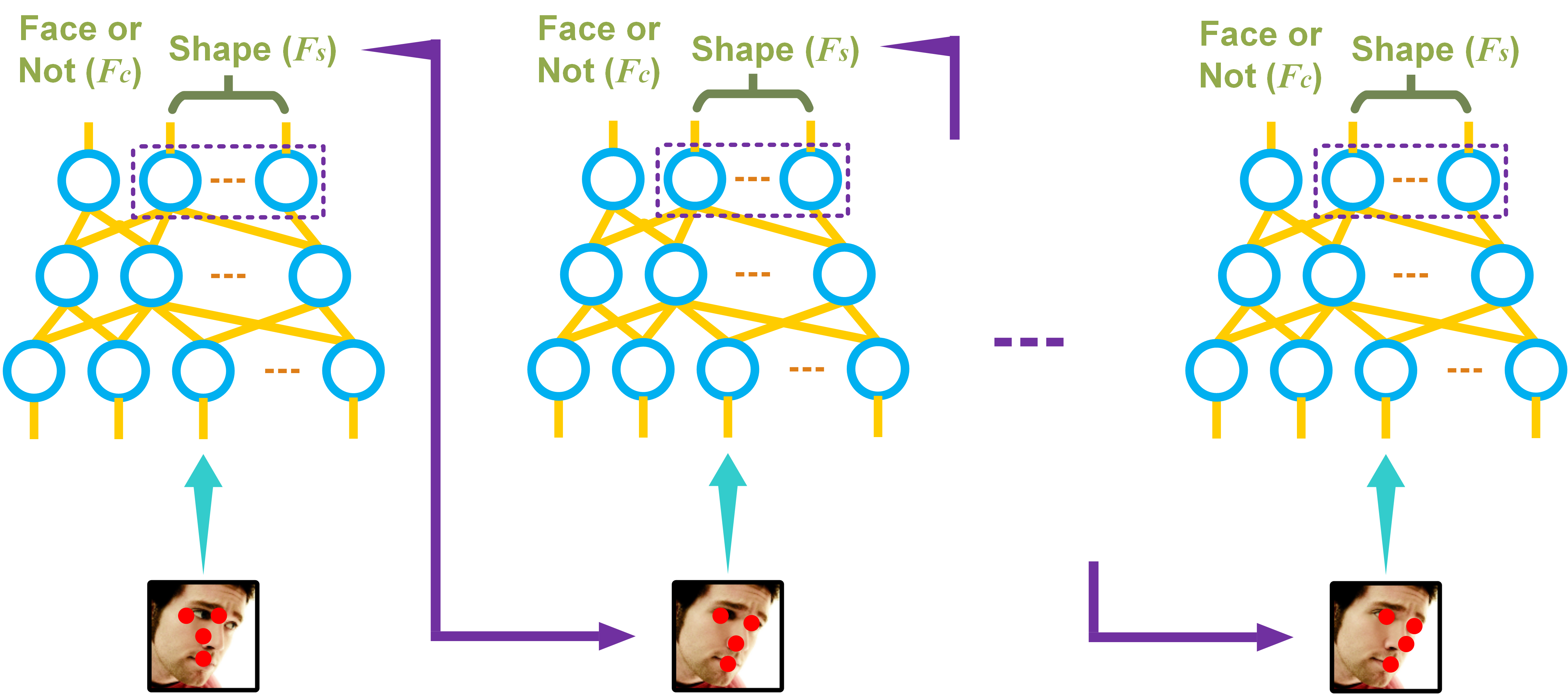}
\end{center}
   \caption{The Fine MLP Cascade with shape-indexed feature. The input of each stage of MLP is the shape-indexed feature extracted according to the shape predicted by the previous stage (or mean shape for the first stage). The output includes the class label indicating whether the window is a face or not as well as a more accurate shape, which is used to extract more distinctive shape-indexed features for the next stage.}
\label{fig:shape_indexed_feature_extraction}
\end{figure*}

Surviving from the previous stages, the small number of windows have been quite challenging, among which face and non-face windows are more difficult to be distinguished. Considering that multiple models running in parallel tend to introduce more false alarms, it is desirable to process the remaining windows in a unified way. Hence we leverage one single MLP cascade following the previous Coarse MLP Cascade classifiers.

\begin{figure}
\begin{center}
   \includegraphics[width=0.65\linewidth]{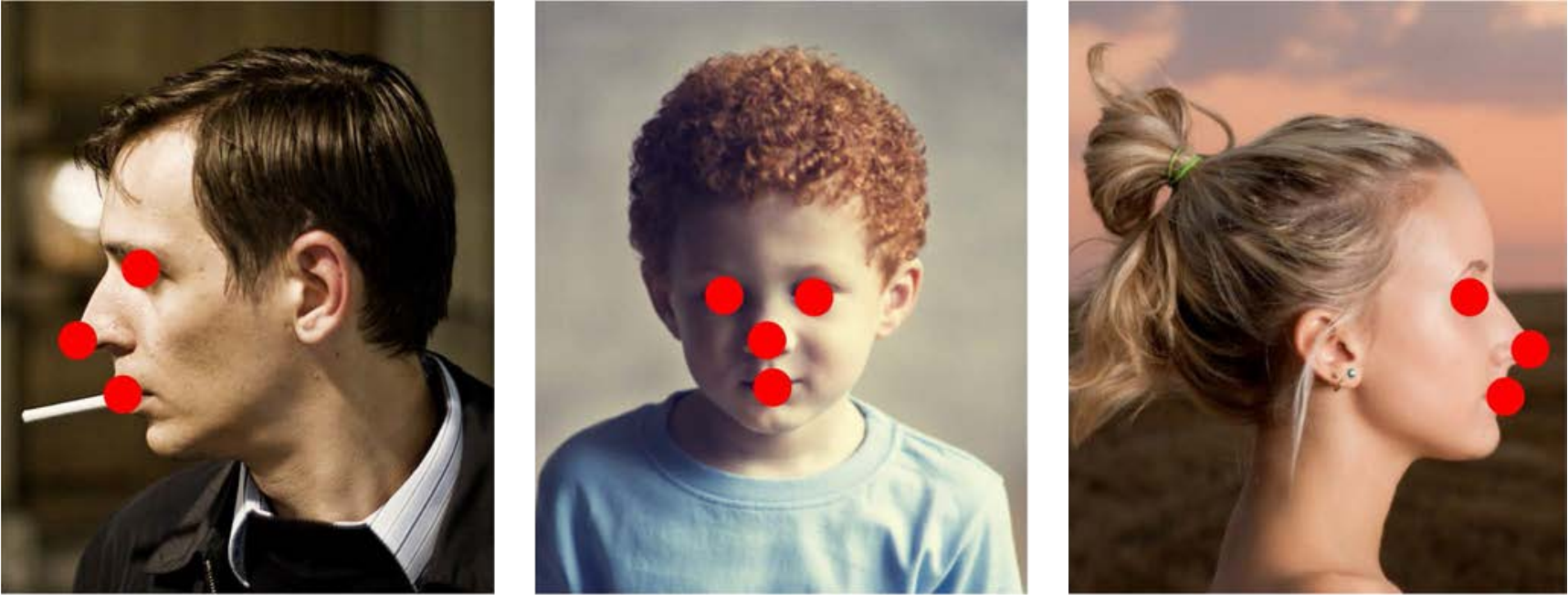}
\end{center}
   \caption{The four semantic positions (landmarks) used to extract shape-indexed feature: left and right eye center, nose tip and mouth center.}
\label{fig:examples_4pts}
\end{figure}

Prominent divergence exists in appearances of multi-view faces, which is mainly due to the unaligned features, i.e. features are extracted at positions that are not semantically consistent. For example, the central region of a frontal face covers the nose, while that of a profile face is part of the cheek, as shown in Figure \ref{fig:faces_not_aligned}. To address this issue, we adopt shape-indexed features extracted at semantically consistent positions as the input of the Fine MLP Cascade classifier. As shown in Figure \ref{fig:examples_4pts}, four semantic positions are selected, corresponding to the facial landmarks of left and right eye center, nose tip and mouth center. For profile faces, the invisible eye is assumed to be at the same position as the other eye. The SIFT (Scale-Invariant Feature Transform) \cite{david2004ijcv} feature is computed at each semantic position on candidate windows, and they are robust to large face variations such as pose, translation, etc.

With the more expressive shape-indexed features, larger MLPs with higher capacity of nonlinearity are used to perform finer discrimination between face and non-face windows. Moreover, different from previous ones, the larger MLPs predict both class label, indicating whether a candidate window is a face, and shape simultaneously. An extra term of shape prediction errors is added to the objective function in Eq. \eqref{eq:mlp_loss}. The new optimization problem is the following
\begin{align} \label{eq:mlp_loss_2}
\min_F \sum_{i=1}^n \! \Vert F_c \! \left( \phi(x_i,\! \hat{s_i}) \right) \! - \! y_i \Vert^2 + \lambda \! \sum_{i = 1}^n \! \Vert F_s \! \left( \phi(x_i,\! \hat{s_i}) \right) \! - \! s_i \Vert_2^2,
\end{align}
where $F_c$ corresponds to the face classification output, and $F_s$ the shape prediction output; $\phi(x_i, \hat{s_i})$ indicates the shape-indexed feature (i.e. SIFT) extracted from the $i$-th training sample $x_i$ according to a mean shape or predicted shape $\hat{s_i}$; $s_i$ is the groundtruth shape for the sample; $\lambda$ is the weighting factor to maintain the balance between the two types of errors, which is set to $\frac{1}{d}$ with $d$ as the dimension of shape. As can be seen from Eq. \eqref{eq:mlp_loss_2}, a more accurate shape $F_s \left( \phi(x_i, \hat{s_i}) \right)$ than the input $\hat{s_i}$ can be obtained with the MLP. Hence a subsequent model can exploit more compact shape-indexed features extracted according to the refined shape $F_s \left( \phi(x_i, \hat{s_i}) \right)$. As so, in multiple cascaded MLPs, the shapes used for feature extraction become more and more accurate stage by stage, leading to more and more distinctive shape-indexed features and further making multi-view faces more distinguishable from non-face regions. The process is shown in Figure \ref{fig:shape_indexed_feature_extraction}.

Additionally, predicting shapes has made the detector alignment-aware in the sense that an alignment model can be initialized with landmark coordinates directly instead of bounding boxes of detected faces.

\section{Experiments}
\label{sec:expt}

To evaluate the proposed FuSt detector for multi-view face detection, as well as to analyse the detector in various aspects, extensive experiments are performed on two challenging face datasets.

\subsection{Experimental settings}

The most popular dataset for evaluating face detectors is the FDDB \cite{vidit2010tr}. It contains $5,171$ labeled faces from $2,845$ news images. FDDB is challenging in the sense that the labeled faces appear with great variations in view, skin color, facial expression, illumination, occlusion, resolution, etc.

Another widely used face detection dataset is the AFW \cite{zhu2012}. This set contains $205$ images from Flickr with $468$ faces. It is a small set, yet is challenging, since faces appears in cluttered backgrounds and with large variations in viewpoints.

For evaluation of the detection accuracy, we apply the officially provided tool to our detection results on FDDB to obtain the ROCs, and draw precision-recall curve for the results on AFW, following most existing works. 

For the training data of the FuSt detector, we use faces from MSRA-CFW \cite{zhang2012tmm}, PubFig \cite{kumar2009iccv}, and AFLW \cite{kostinger2015iccvw} as positive samples, and randomly crop patches from $40,000$ collected images not containing faces as negative samples. To augment the training set with more variations, we add random distortions to the face samples. Besides, all samples are resized to $40\times 40$ for training.

We use $1$ stage with a total of $150$ LAB features for the Fast LAB Cascade, and $3$ stages for the Coarse MLP Cascade, which exploit $2$, $4$ and $6$ SURF features respectively. SURF features are extracted based on local patches, which will cover redundant information if there is considerable overlap between them. Therefore a large step of $16$ are chosen for adjacent SURF patches, resulting in a pool of $56$ SURF features on a $40\times 40$ sample image. The three stages of MLP all have only one hidden layer, and there are $15$ hidden units in the first-stage MLP and $20$ hidden units in the second- and third-stage MLP. The final Fine MLP Cascade contains $2$ stages of single-hidden-layer MLP with $80$ hidden units with SIFT features extracted around the four semantic positions as mentioned in Section \ref{ssec:fine_mlp_cas}.

\subsection{Analysis of the funnel-structured cascade}

\label{ssec:expt_analysis}

We first conduct a detailed analysis of the proposed FuSt detector to evaluate its performance from various perspectives. Specifically, we compare different view partitions, verify the effectiveness of shape-indexed features, assess the accuracy of shape predictions, and compare the final MLP cascade with two widely used CNN models.

\paragraph{Different view partitions} At the beginning, we adopt a divide-and-conquer strategy to treat faces in different views with separate LAB cascade classifiers. This makes it possible for such simple classifiers to reject a large proportion of non-faces windows, while retaining a high overall recall of faces. To explore the impact of different view partitions, we compare two typical partition schemes: (1) five-view partition, i.e. \textit{left full profile, left half profile, near frontal, right half profile, and right full profile}; (2) two-view partition, i.e. \textit{near frontal, profile}. Note that in the second two-view partition scheme, left and right profile faces are mixed together, and half profile faces are mixed with frontal ones. To supplement the training set with more half profile face images, we also use some images from CelebA dataset \cite{liu2015iccv}. The recall of faces with the two schemes are presented in Table \ref{table:view_partition_recall}. Here we manually partition the FDDB into two subsets of profile and frontal faces to evaluate on them separately. The former contains $527$ profile faces from $428$ images, and the latter, i.e. the frontal face subset, contains the rest faces including both near frontal and some half profile faces.

\begin{table}
\begin{center}
\begin{tabular}{|c|ccc|}
\hline
View & \multicolumn{3}{c|}{Recall of Faces (\%)} \\ 
\cline{2-4}
Partition & Frontal & Profile & Overall \\
\hline\hline
$5$ Views & $96.27$ & $95.83$ & $96.15$ \\
$2$ Views & $95.07$ & $92.60$ & $94.82$ \\
\hline
\end{tabular}
\end{center}
\vspace{-10pt}
\caption{Recall of faces with different view partitions with over $99\%$ windows removed}
\label{table:view_partition_recall}
\end{table}

As can be seen, the recall of faces with the five-view partition, especially the recall of profile faces, are higher than that with the two-view partition when both scheme remove over $99\%$ of candidate windows. As expected, the finer partition allows classifiers to cover more variations within each view of faces, and is beneficial for obtaining higher recall. This demonstrates the effectiveness of using a reasonably wide top in the proposed funnel structure.

\paragraph{Funnel structure vs parallel structure} To demonstrate the effectiveness of the proposed funnel structure employing a unified model to handle candidate windows coming from different classifiers, we compare the parallel and the funnel structure on frontal and half profile faces in the coarse MLP cascade stage. Specifically, for the parallel structure, we train three MLPs, one for each of the three views, which follows the corresponding fast LAB cascade. For the funnel structure, only one MLP is trained for frontal, left half profile and right half profile faces. The parallel structure obtains a recall of $94.41\%$ with $297.06$ windows per image, while the funnel structure reaches a higher recall of $94.43\%$ with only $268.10$ windows per image. This demonstrates that a unified model can effectively control the false positives with less sacrifice of recall.

\paragraph{Shape-indexed feature} To verify the effectiveness of the shape-indexed feature, we train two types of two-stage Fine MLP Cascade classifiers with mean shape and refined shape respectively, which are used to extract shape-indexed feature. Namely, one MLP cascade uses SIFT extracted according to mean shape as input at both stages, while the other uses SIFT extracted with refined and thus more accurate shapes as input at the second stage.

\begin{figure}
\begin{center}
   \includegraphics[width=0.9\linewidth]{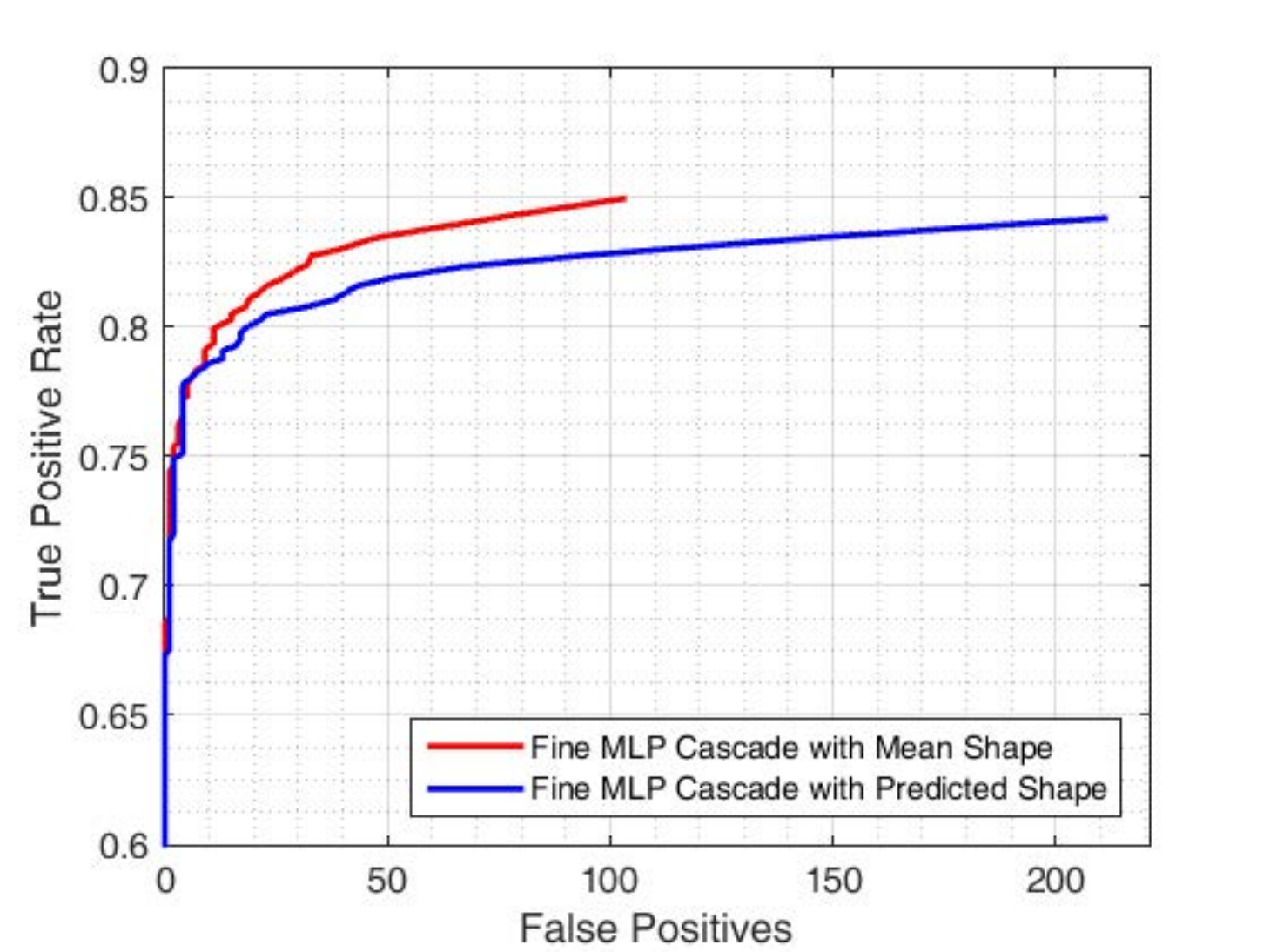}
\end{center}
   \caption{Comparison between shape-indexed features extracted with mean shape and refined shape}
   \label{fig:cmp_shape_indexed_feature}
\end{figure}

Fixing previous stages, we compare the two types of Fine MLP Cascades on FDDB. The performance curves are presented in Figure \ref{fig:cmp_shape_indexed_feature}. As expected, using more accurate shapes brings performance gain, demonstrating the effectiveness of shape-indexed features for multi-view faces. Shape-indexed features from two faces have good semantic consistence, thus reducing intra-class variations and increasing inter-class distinctions. This makes it easier to distinguish face from non-face windows.

\begin{figure}
\begin{center}
   \includegraphics[width=0.9\linewidth]{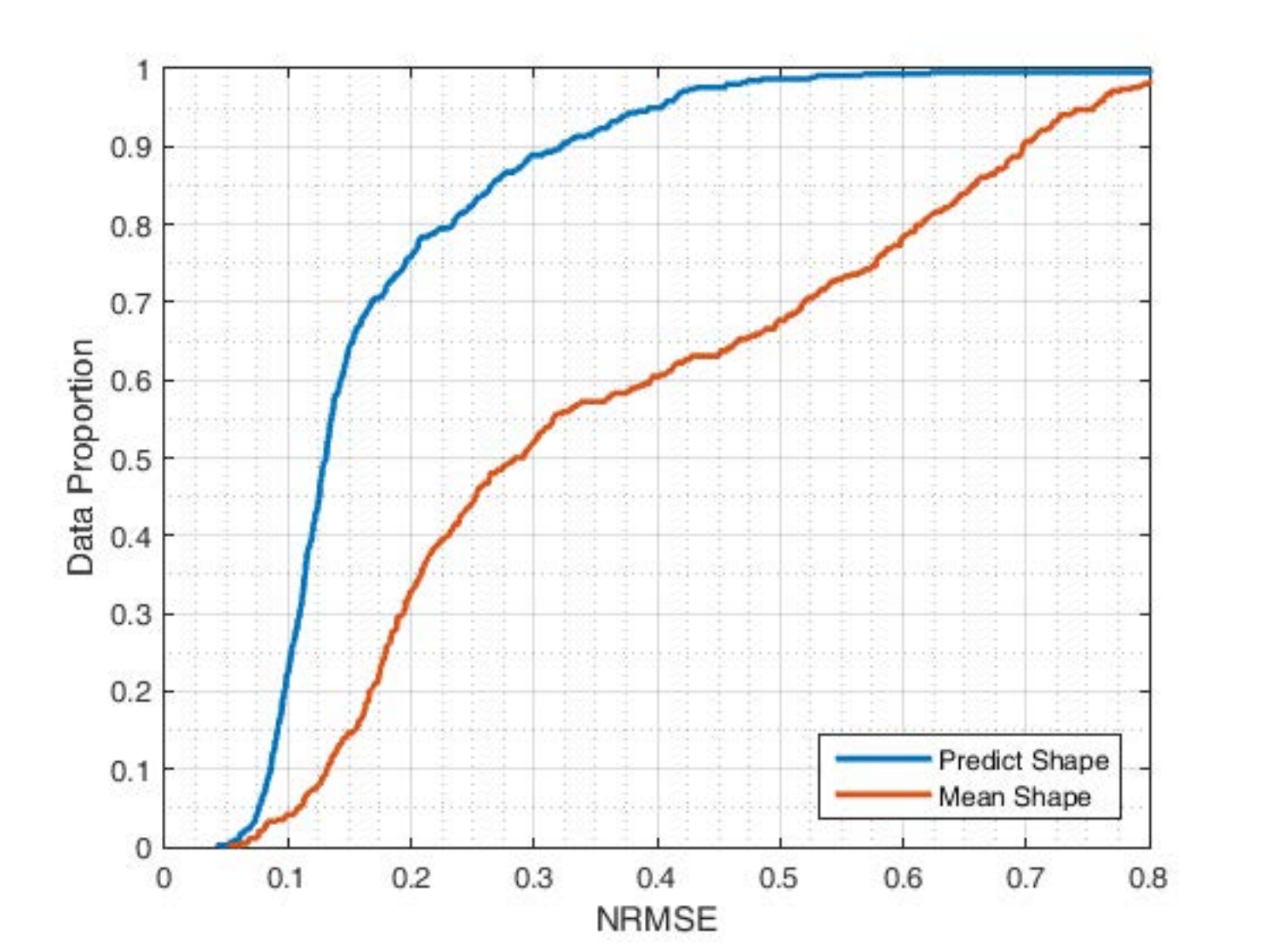}
\end{center}
   \caption{Comparison between predicted shape and mean shape on AFW}
   \label{fig:cmp_afw_shape_predict_accuracy}
\end{figure}

\begin{figure}
\begin{center}
   \includegraphics[width=0.9\linewidth]{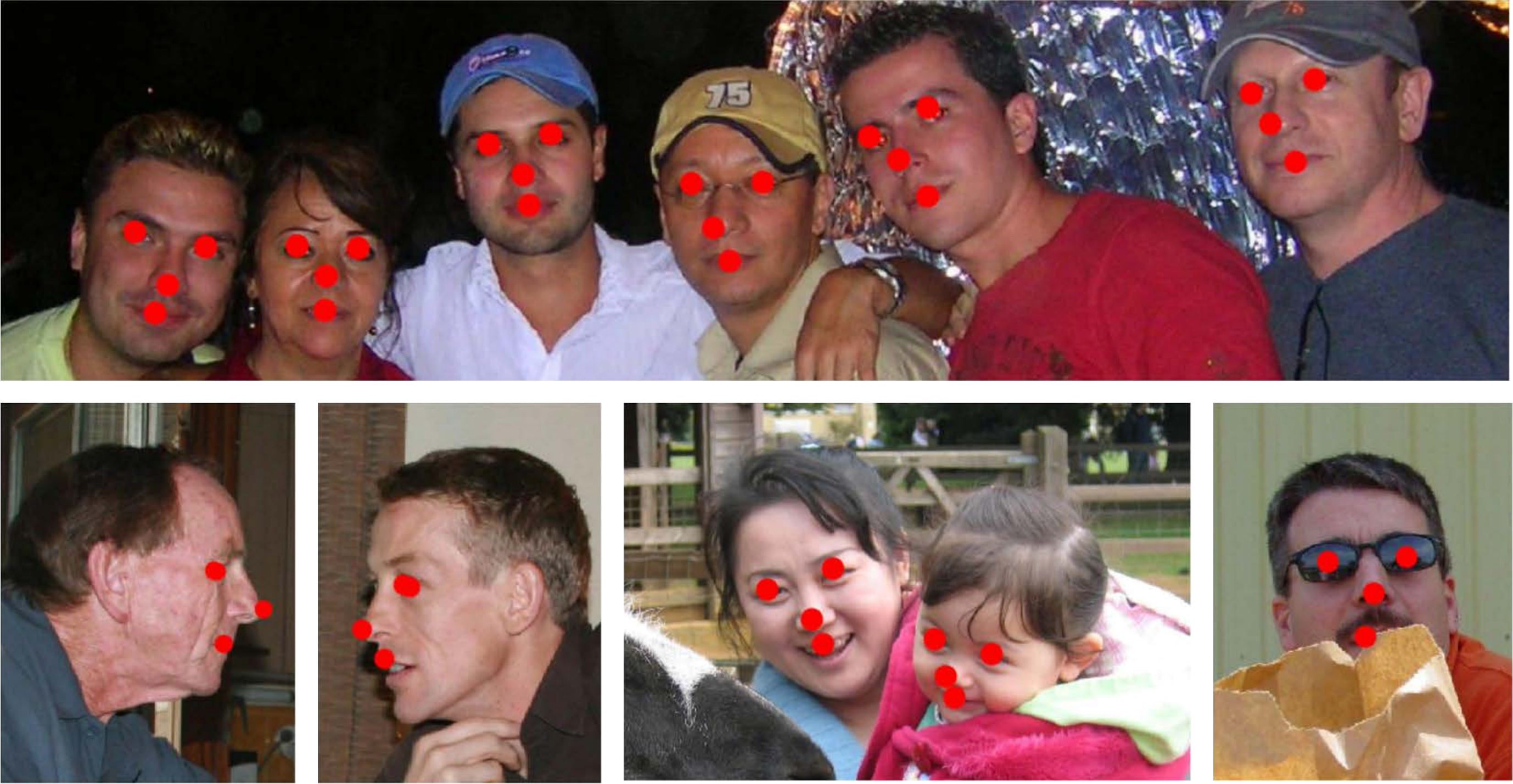}
\end{center}
   \caption{Examples of predicted shapes on AFW}
   \label{fig:afw_shape_predict_examples}
\end{figure}

We also evaluate the coarse shape predictions on AFW. Figure \ref{fig:cmp_afw_shape_predict_accuracy} compares the predicted shape with the mean shape. With only two stages of refinement, the predicted shapes achieve significant improvement over the mean shape, leading to more semantically consistent shape-indexed features. When followed by an alignment model, the predicted shape from our FuSt detector can be directly used as a good initialization, which is more preferable than only bounding boxes of detected faces. Figure \ref{fig:afw_shape_predict_examples} gives several examples of predicted shapes on faces in different views.

\begin{figure}
\begin{center}
   \includegraphics[width=0.9\linewidth]{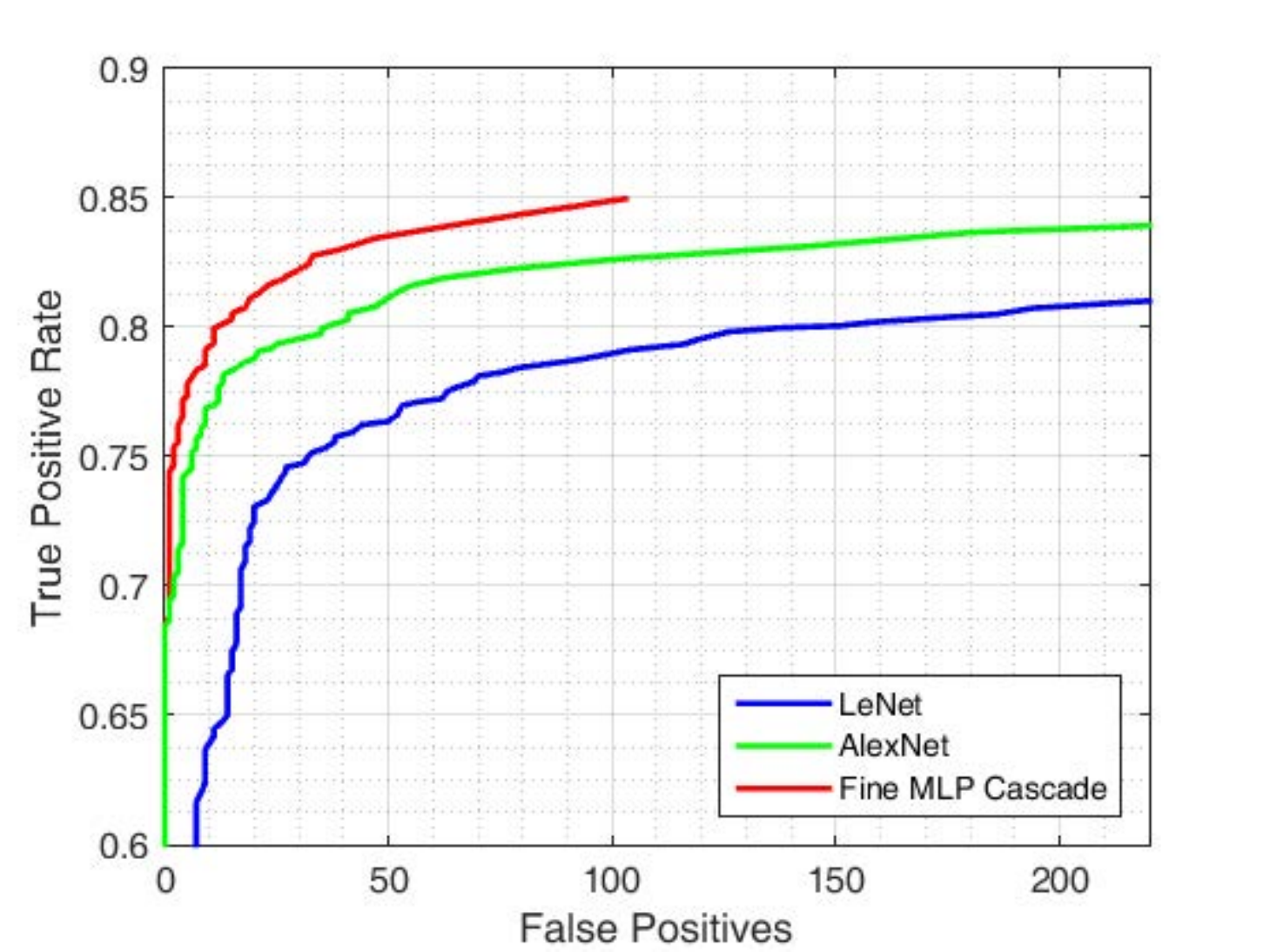}
\end{center}
   \caption{Comparison of MLP cascade, LeNet and AlexNet}
   \label{fig:cmp_mlp_lenet_alexnet}
\end{figure}

\paragraph{MLP vs CNN} The powerful CNN models have achieved good results in face detection task \cite{farfade2015icmr, li2015cvpr, yang2015iccv:faceness}, so we also compare MLP with CNN under the proposed funnel-structured cascade framework. Two commonly used CNN models are considered in the comparison, i.e. LeNet \cite{lecun1998ieee} and AlexNet \cite{alex2012nips}, and they serve as replacements for the final Fine MLP Cascade. The input sizes of LeNet and AlexNet are $40\times 40 \times 3$ and $256\times 256 \times 3$ respectively, and the output layers are adjusted for two-class classification of face or non-face. Both CNN models are fine-tuned using the same data as that used in training the MLP cascade. The performance curves on FDDB are given in Figure \ref{fig:cmp_mlp_lenet_alexnet}. As is shown, the MLP cascade outperforms LeNet by a large margin and also performs better than the $8$-layer AlexNet. This is most likely because the semantically consistent shape-indexed features are more effective than the learned convolutional features. Considering the result that the MLP with hand-crafted features has the ability to defeat deep CNN models, it implies that a well-designed model with considerations for the problem can be better than an off-the-shelf CNN.

\paragraph{Detection Speed} Our FuSt detector enjoys a good advantage of detection speed with the coarse-to-fine framework design and is faster than complex CNN-based detectors. When detecting faces no smaller than $80\times 80$ on a VGA image of size $640\times 480$, our detector takes $50$ms with step-$1$ sliding window using a single thread on an i7 CPU. The Fast LAB Cascade and Coarse MLP Cascade cost only $30$ms, and the final Fine MLP Cascade $20$ms. By contrast, Cascade CNN takes $110$ms over an image pyramid with scaling factor of $1.414$ on CPU \cite{li2015cvpr}. Moreover, further speed-up of FuSt detector can be easily obtained with GPU since a large amount of data parallelism exists in our framework, e.g. feature extraction for each window, the inner product operations in MLP, etc.

\begin{table}
\begin{center}\small
\begin{tabular}{|c|ccc|}
\hline
Methods & DR@100FPs & Speed & \begin{tabular}[x]{@{}c@{}}Landmark\\Prediction\end{tabular}  \\
\hline\hline
Cascade CNN \cite{li2015cvpr} & $85\%$ & $110$ms & No \\
Our FuSt & $85\%$ & $50$ms & Yes \\
\hline
\end{tabular}
\end{center}
\caption{Comparison with Cascade CNN \cite{li2015cvpr} in different aspects. The DR@100FPs is computed on FDDB, and the speed is compared with minimum face size set as $80\times 80$ and image size $640\times 480$.}
\label{table:comparison_cascadecnn}
\end{table}

\paragraph{Discussion} Compared with CNN based methods, the proposed funnel structure is a general framework of organizing multiple models, adopting a divide-and-conquer strategy to handle multi-view faces. The MLPs used with the framework can also be replaced by CNNs. One other aspect that makes our FuSt detector different is that hand-crafted shape-indexed feature is adopted based on explicit consideration for semantically consistent feature representation. By contrast, CNN learns the feature representation merely from data without considering the semantic consistency.

\subsection{Comparison with the state-of-the-art}

\begin{figure}
\begin{center}
    \begin{subfigure}[t]{0.9\linewidth}
        \centering
        \includegraphics[width=\linewidth]{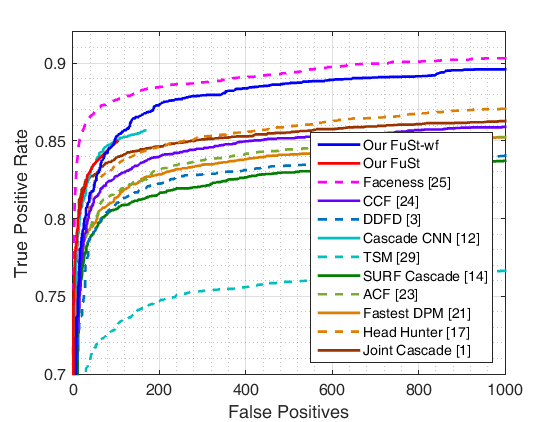}
        \caption{FDDB}
        \label{fig:cmp_fddb_2}
    \end{subfigure}
    
    \begin{subfigure}[t]{0.9\linewidth}
        \centering
        \includegraphics[width=\linewidth]{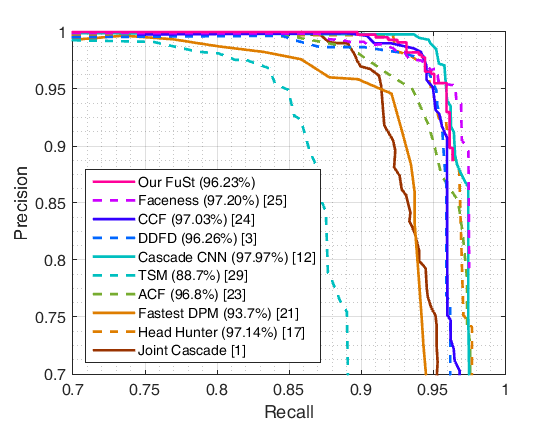}
        \caption{AFW}
        \label{fig:cmp_afw}
    \end{subfigure}
\end{center}
   \caption{Comparison with the state-of-the-art on two face detection datasets: (a) FDDB, (b) AFW.}
\label{fig:cmp_state_of_the_art}
\end{figure}

To further evaluate the performance of the FuSt detector on multi-view face detection, we compare it with the state-of-the-art methods on FDDB and AFW as shown in Figure \ref{fig:cmp_state_of_the_art}. Methods being compared include cascade-structured detectors such as Joint Cascade \cite{chen2014}, ACF \cite{yang2014}, SURF Cascade \cite{li2013}, and Head Hunter \cite{mathias2014}, DPM-based detectors such as Fastest DPM \cite{yan2014}, and TSM \cite{zhu2012}, and deep-network-based detectors such as DDFD \cite{farfade2015icmr}, Cascade CNN \cite{li2015cvpr}, CCF \cite{yang2015iccv:ccf}, and FacenessNet \cite{yang2015iccv:faceness}.

Compared with multi-view face detectors like SURF Cascade, ACF, and Head Hunter, which all employ a parallel structure, our FuSt detector performs better on FDDB, indicating the superiority of our funnel structure. With as few as $100$ false positives, the FuSt detector achieves a high recall of $85\%$, which is quite favorable in practical applications. Compared with the impressive deep-network-based methods, we achieve comparable performance with that of Cascade CNN. However, as stated in Section \ref{ssec:expt_analysis}, our FuSt detector enjoys a more favorable speed, taking only $50$ms to detect a VGA image with a single thread on CPU. By contrast, Cascade CNN costs $110$ms on CPU. On AFW dataset, our PR curve is comparable to or better than most methods, further demonstrating that our FuSt detector is favorable for multi-view face detection.

To further investigate the potential of our FuSt detector on FDDB, we trained a new detector FuSt-wf with a more diverse dataset WIDER FACE \cite{yang2016cvpr}. WIDER FACE dataset covers much more face variations, which is beneficial for obtaining higher performance. Since WIDER FACE does not provide landmark annotations for faces, we only trained one stage for the unified MLP cascade with mean shape. As shown in Figure \ref{fig:cmp_state_of_the_art}, FuSt-wf achieves obvious performance boost, further demonstrating the effectiveness of the funnel-structure design. With higher quality and more data, the FuSt detector can continue to improve.

\begin{figure*}
\begin{center}
   \includegraphics[width=0.85\linewidth]{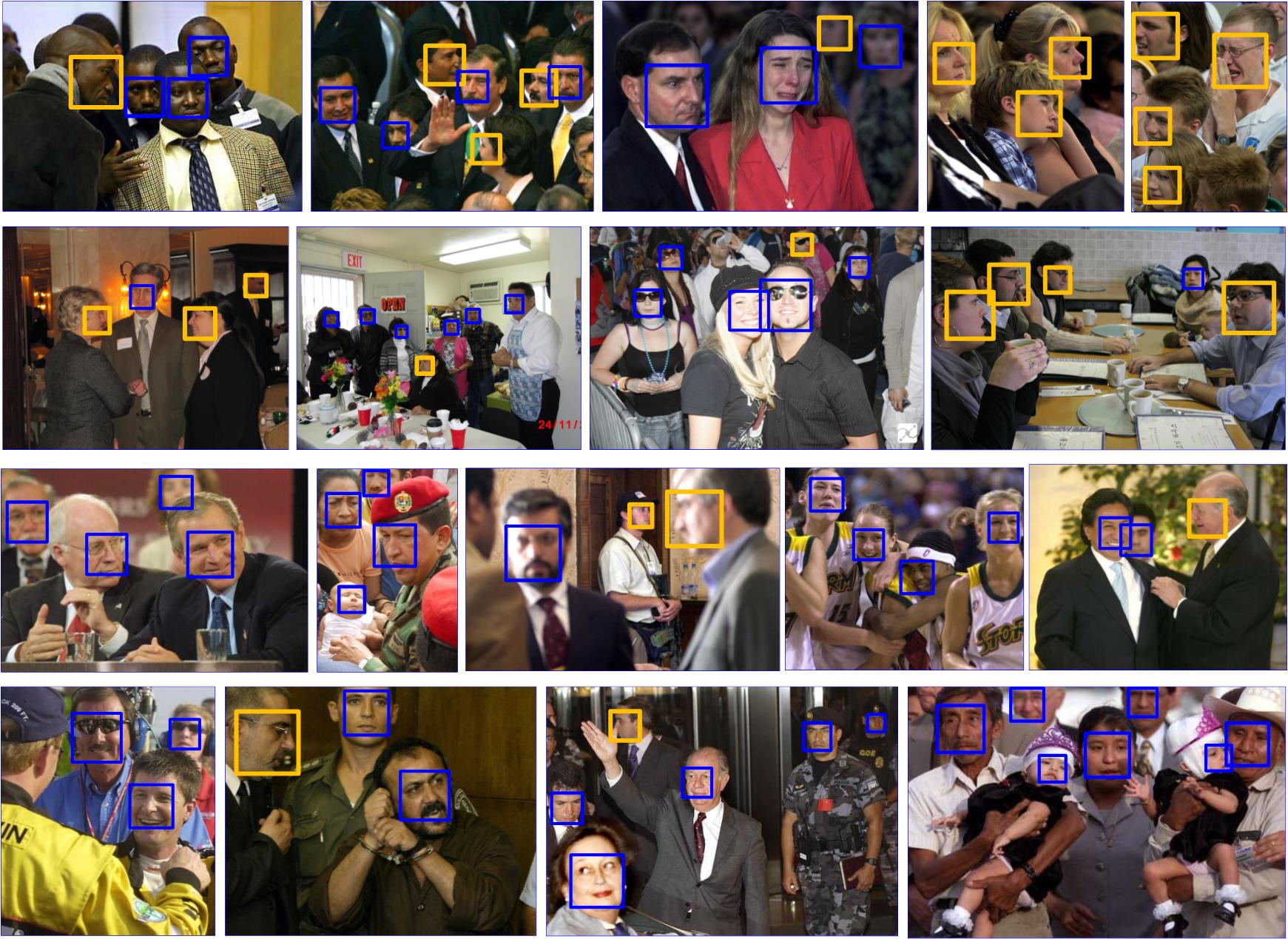}
\end{center}
   \caption{Examples of detections on FDDB and AFW (Blue: near frontal faces, Orange: profile faces)}
\label{fig:detection_result}
\end{figure*}

\section{Conclusions and Future Works}
\label{sec:conclude}

In this paper, we have proposed a novel multi-view face detection framework, i.e. the funnel-structured cascade (FuSt), which has a coarse-to-fine flavor and is alignment-aware. The proposed FuSt detector operates in a gathering style, with the early stages of multiple parallel models reaching a high recall of faces at low cost and the final unified MLP cascade well reducing false alarms. As evaluated on two challenging datasets, the FuSt detector has shown good performance, and the speed of the detector is also quite favorable. In addition, the alignment-awareness nature of our FuSt detector can be leveraged to achieve a good initial shape for subsequent alignment models with minor cost.

For the future work, the funnel structure framework can be further enhanced with specifically designed CNN models which have good capability of learning feature representations automatically from data. It is also worth trying different hand-crafted shape-indexed features, e.g. the multi-scale pixel difference features used in \cite{chen2014}, and comparing them with CNN-learned features. Considering the alignment-awareness nature of the FuSt detector, it is also a promising direction to design a joint face detection and alignment framework.

\section*{Acknowledgements}

This work was partially supported by 973 Program under contract No. 2015CB351802, Natural Science Foundation of China under contracts Nos. 61173065, 61222211, 61402443 and 61390511.

{\small
\bibliographystyle{ieee}
\bibliography{egbib}
}

\end{document}